\documentclass[letterpaper, 10 pt, conference]{ieeeconf}  

\IEEEoverridecommandlockouts                              

\usepackage{graphics} 
\usepackage{subfig}
\usepackage{epsfig} 
\usepackage{dblfloatfix}
\usepackage{pdflscape}

\usepackage[latin1]{inputenc} 
\usepackage[T1]{fontenc} 
\usepackage[ngerman]{babel}

\title{\LARGE \bf
Reasoning Systems for Semantic Navigation in Mobile Robots 
\thanks{ {\bf This is the authors' manuscript. The final published article is
available at https://doi.org/10.1109/IROS.2018.8594271}}
\thanks{ {\bf Copyright 2018 IEEE. Personal use of this material is permitted. Permission
from IEEE must be obtained for all other uses, in any current or future
media, including reprinting/republishing this material for advertising
or promotional purposes, creating new collective works, for resale or
redistribution to servers or lists, or reuse of any copyrighted
component of this work in other works.}}
}

\author{Jonathan Crespo$^{1,*}$, Ram\'on Barber$^{1}$, O. M. Mozos$^{2}$, Daniel Be{\ss}ler$^{3}$ and Michael Beetz$^{3}$
\thanks{$^{1}$Jonathan Crespo ({\tt\small jocrespo@ing.uc3m.es}) and Ramon Barber are with University of Carlos III of Madrid, 28911, Spain.
	}%
\thanks{$^{2}$Oscar M. Mozos is with Technical University of Cartagena, 30202, Spain.
 }%
\thanks{$^{3}$Daniel Be{\ss}ler and Michael Beetz are with Institute of Artificial Intelligent at the University of Bremen, 28359, Germany.
	}%
\thanks{$^{*}$ Corresponding author.}%
}

\begin{document}

\maketitle
\thispagestyle{empty}
\pagestyle{empty}

\begin{abstract}

Semantic navigation is the navigation paradigm in which environmental semantic concepts and their relationships are taken into account to plan the route of a mobile robot. This paradigm facilitates the interaction with humans and the understanding of human environments in terms of navigation goals and tasks. At the high level, a semantic navigation system requires two main components: a semantic representation of the environment, and a reasoner system. This paper is focused on develop a model of the environment using semantic concepts. This paper presents two solutions for the semantic navigation paradigm. Both systems implement an ontological model. Whilst the first one uses a relational database, the second one is based on KnowRob. Both systems have been integrated in a semantic navigator. We compare both systems at the qualitative and quantitative levels, and present an implementation on a mobile robot as a proof of concept.
\end{abstract}

\section{INTRODUCTION}

During the past years there is a growing interest in the provision of high-level cognitive skills in all robotics applications. Robots that navigate in human environments are an example of an application that can benefit from high level knowledge. Service robots which work in human environments need high level perception and behavioral mechanisms closer to human models of perception and reasoning. Ioannis Kostavelis et al. ensure in \cite{Kostavelis2016}. This is also commented in \cite{Crespo2017}.

The navigation task requires the robot to build a model of the environment. This requirement is met by robots that navigate, from robotic wheelchairs \cite{Nguyen2018} to marine surface vehicles \cite{Sun2018}. Of course, mobile robots that navigate in human environments are included. In this field, some authors are currently focusing on modeling the environment where the world consists only of obstacles and free spaces \cite{Nelson2018}. However, there is increasing interest in using a higher level of modeling abstraction for robots in indoor environments. A higher level of abstraction is a decisive change in the abilities of the robot. This gives the robot more autonomy, more robustness and more efficiency.

Increasing the level of abstraction implies that the robot must be able to handle more complex concepts of the environment. Modeling the world is no longer limited to representing obstacles and hierarchical maps of concepts arise as in \cite{Hao2015}. Concepts related to concrete objects or the actions that can be performed with those objects are included. The objects are related to room types. These concepts are part of the environment. Semantic navigation is the navigation system that takes into account the information of these concepts to model the environment. The so-called semantic maps are used to improve planning methods \cite{Drouilly2015}.

Semantic navigation also requires increasing other capabilities of the robot. These navigation systems usually require that an ontology is designed \cite{Paull2012}. Ontologies are one of the most widely used ways of representing information and relationships of concepts \cite{Schlenoff2012}. The use of ontologies expedites the hierarchies of concepts definition. Some authors take advantage of this technology to define hierarchies with different levels of abstraction \cite{H.Zender2008}. However, the management of the ontology needs a reasoning system to be added. Different technologies for the robot reasoning are in the literature. Each applied technology requires a different way of implementing an ontology. The system proposed in \cite{C.Galindo2005} is a common example of implementation. The ontology is defined with a set of rules. The rules are manually coded by researchers in a reasoning system called NeoClassic. This system is based on descriptive logics, such as LOOM \cite{C.Galindo2008}, and the system makes the inferences.

In this work, two technologies to implement the ontology have been implemented and studied. Each technology has needed a different system to make the inferences. First, an environment has been defined for a navigation experiment with a mobile robot. Afterwards, the planning system has been asked for different objectives. The planner identifies the place to which the robot has to go. For this, the system needs to reason about the objective. This reasoning has been done with the two implementations. On the one hand, a reasoner has been implemented using KnowRob \cite{Tenorth2013} as a reasoning system. This system uses an ontology defined in an OWL (Web Ontology Language) file generated with Prot\'eg\'e. Prot\'eg\'e is an ontology and knowledge base editor produced by Stanford University. Prot\'eg\'e is a tool that enables the construction of domain ontologies. This tool is used in many works like \cite{Chatterjee2018} and \cite{Satyamurty2018}. On the other hand, the ontology has been implemented in a relational database. In addition, software has been developed to extract inferences with the information contained in the database. The database system is published in \cite{Crespo2017}.

This work aims to discuss the characteristics of each technology.

\section{KNOWLEDGE REPRESENTATIONS FOR SEMANTIC NAVIGATION INDOORS}

\begin{figure}[ht!]\centering
	\includegraphics[width=\columnwidth]{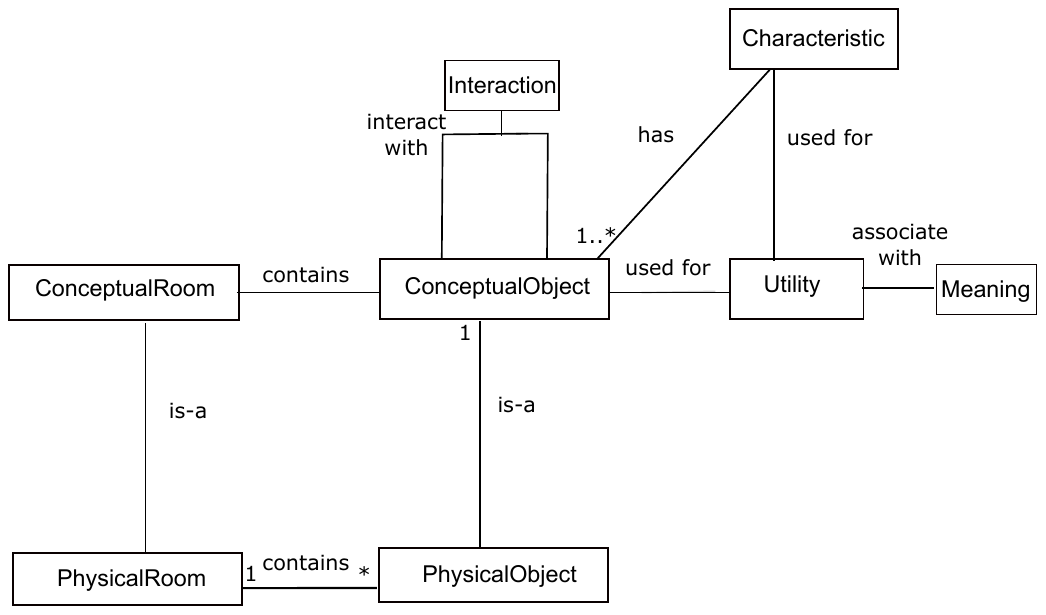}\\
	\caption{Ontology model entity-relationship diagram}
	\label{dontologico}
\end{figure}

Semantic modeling is used to represent knowledge in a relational model. The entity-relationship diagram used is shown in figure \ref{dontologico}. Conceptual objects and physical rooms are related to physical objects and physical rooms. This is supported by ideas proposed by other authors such as H. Zender et al. \cite {H.Zender2008}. They unite the conceptual hierarchy with the spatial hierarchy.

The objects in domestic environments have been taken into account to design the ontology. This is supported by other knowledge models such as the one used in \cite{Jung2007}. Jung et al. relate objects with verbs by means of the relation \textit{usedFor}. It represents the same as the relation \textit{utility} used in this work.

The model used in this paper (Figure \ref{dontologico}) collects the main aspects of the proposed environment and its semantic representation. The system must manage the following four elements: \textit{Conceptual Room} (room abstract concept), \textit{Conceptual Object} (object abstract concept), \textit{Physical Room} (identifiable room in the real world) and \textit{Physical Object} (identifiable object in the real world). These entities are complemented with the following concepts: \textit{Interaction} (usually objects interact with other objects, such as the way they are used), \textit{utility} and \textit{meaning} (a subjective appreciation of what some actions bring). This all concepts are described on \cite{J.Crespo2015}.

The concept \textit{utility} is a concept that describes the actions that can be performed with an object. This is important for navigation because the objects that are used for tasks that are related are usually in the same place. The \textit{meaning} concept is a concept that relates a \textit{utility} to a subjective feeling that the action produces in the user. For example, the \textit{play} action produces a feeling that is identified with \textit{funny}. Sitting on a sofa produces a feeling that is \textit{relaxing}.

 In the navigation system, the reasoner that manages these concepts plays an important role. The generic reasoner offers methods for making inferences used in semantic navigator. The methods that the reasoner class has got can be separated into two groups and are the methods that extract stored knowledge. On the other hand, are the methods that infer information based on the data that the robot currently possesses (both stored knowledge and current sensory perceptions). In this paper, two reasoners based on different technologies are used: a model based on ontologies mechanism and a model based on relational models. Both specific reasoners extend a generic reasoner class.

\section{SQL-BASED SYSTEM}

\begin{figure}[ht!]
	\centering
	\includegraphics[width=\columnwidth]{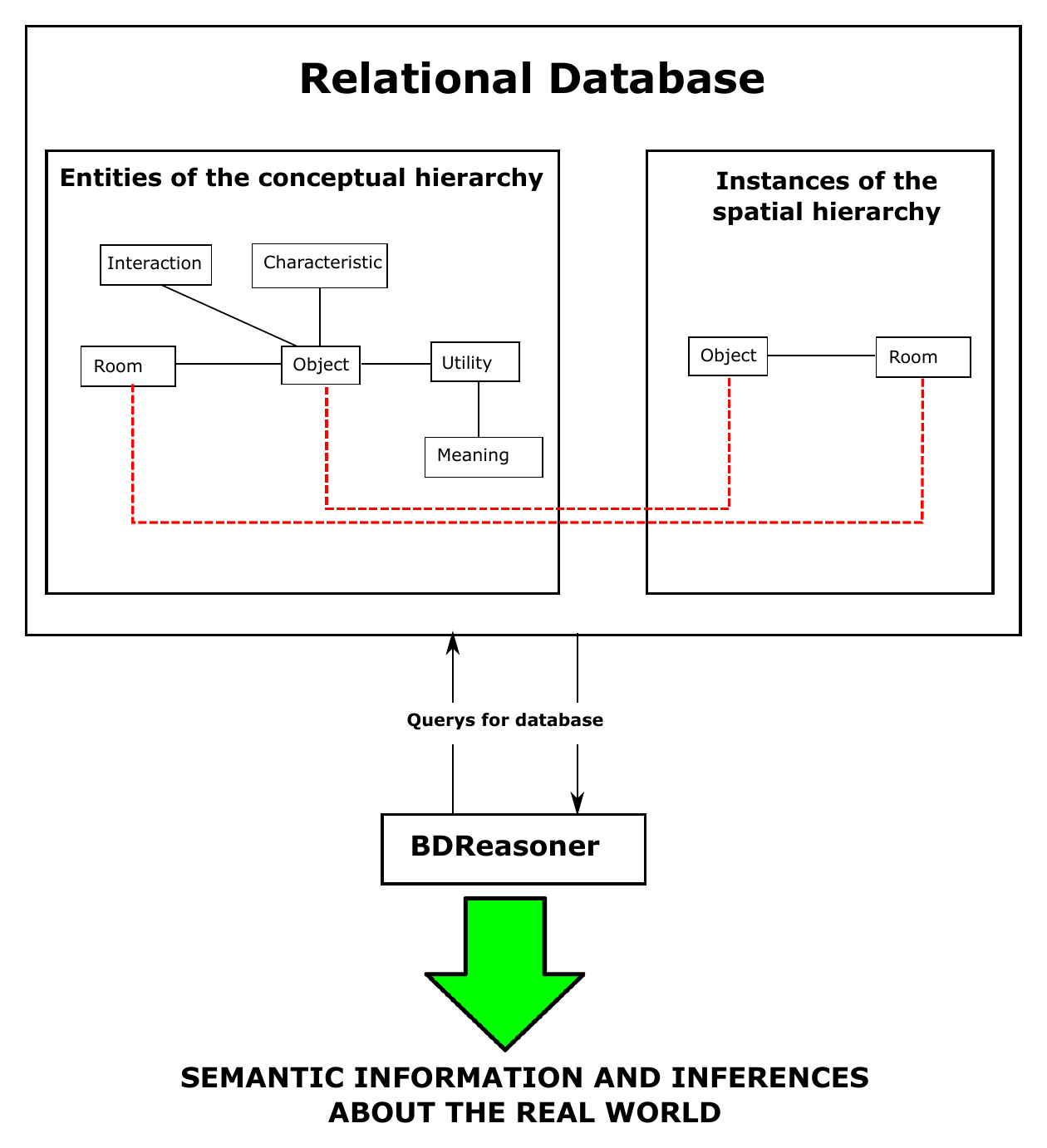}\\
	\caption{Extracting information from a relational data base.}
	\label{resuRazonadorBD}
\end{figure}

The first reasoning system used in this work has been described and tested in previous papers \cite{J.Crespo2015} y \cite{Crespo2017}. It is based on a relational database. The relations between the semantic concepts are replicated on the relationship between the entities of the database. The basic concept of the reasoning process is based on the premise that if a relational database contains both the entities of the conceptual hierarchy and the instances of the physical hierarchy. And if that information is stored in lists and those lists are related to each other as in the entity-relationship model of the environment. Then  the information needed by a reasoning system oriented to semantic navigation can be obtained through a series of predefined SQL queries. This idea is represented graphically in the figure \ref{resuRazonadorBD}.

\begin{figure}[ht!]
	\centering
	\includegraphics[width=\columnwidth]{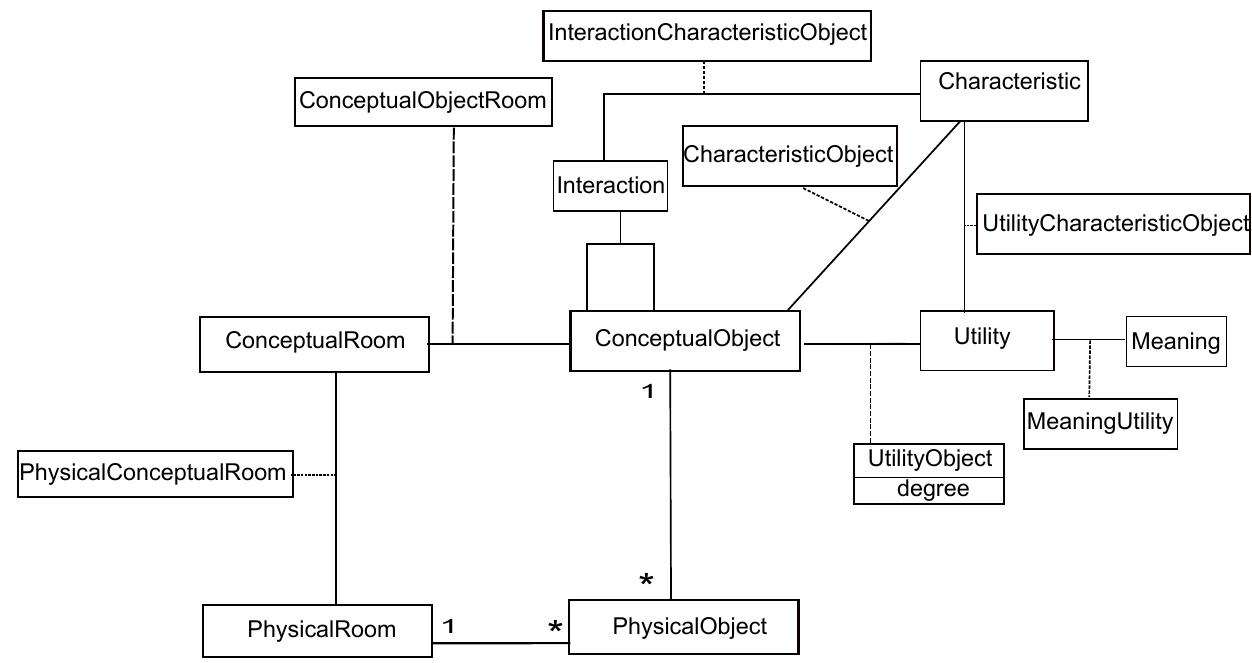}\\
	\caption{Relational model representation of the SQL based system reasoner.}
	\label{tablasBD}
\end{figure}

The ontology entity-relationship diagram in Figure \ref{dontologico} is implemented in SQL tables (Figure \ref{tablasBD}). This implementation generates a table for each concept and tables for many-to-many relationships are also added. All relationships between concepts are implemented as relationships between tables in the database. For example, the meaning table is related to the utility table with a many-to-many relationship. This causes a new table to emerge. This new table has the identifier of the action and the identifier of the meaning. In this way the actions, the meanings and their relationships can be stored.

\section{KNOWROB-BASED SYSTEM}

\begin{figure*} [ht!]
	\centering
	\includegraphics[width=0.6\textwidth]{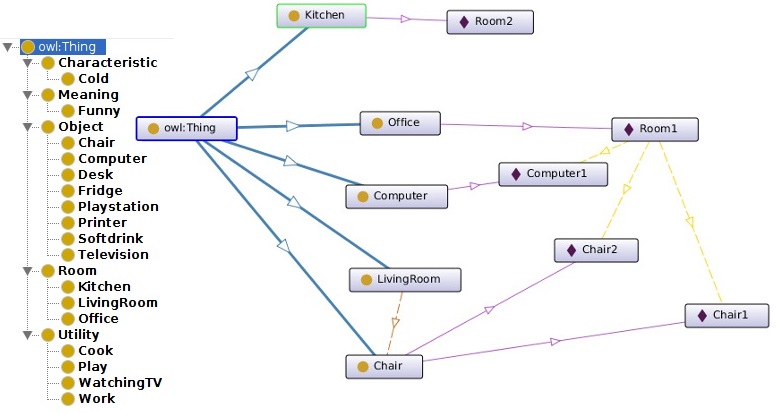}\\
	\caption{Relationships of the conceptual and physical hierarchy implemented in Protege.}
	\label{ontoKR3}
\end{figure*}

Knowrob works as a common semantic framework for integrating information from different sources. The ontology is defined with the Protege software, described in \cite{Noy2003}. Protege generates a .owl file. The file is subsequently loaded by a ROS (Robotic operating system used in the implementation) server that attends the reasoner's queries. The first step is to migrate the ontology to the new software. Figure \ref{ontoKR3} shows several concepts of the conceptual hierarchy. The constructed tree is determined by \textit{is-a} relations. ``Thing'' is the original concept and others inherit from it. The first level is formed by the entities \textit{Characteristic}, \textit{Meaning}, \textit{Object}, \textit{Room} and \textit{Utility}. ``Cold'' has been introduced as a characteristic, ``Funny'' as a meaning. In addition three types of room and four utilities have been introduced.


However, introducing restrictions and other types of relationships that are modeled as restrictions is necessary to represent the relationships between these entities conceptually. In this way, the ``Cold'' characteristic is associated with the \textit{Refrigerator} and all the contained objects are related with ``Cold''. For this it is necessary to establish the relation \textit{contain}, which is also used to specify which objects are contained in the different types of room. Objects are also related to their utility and utilities are related to subjective meanings. For example, television is related to the ``watching television'' utility  and the ``watching television'' utility is related to the subjective meaning \textit{funny}.

These relationships are shown in Figure \ref{ontoKR3}, which shows the complete ontology that has been used in the examples to perform the first tests on this reasoner. The relationships between concepts have been implemented by establishing properties. This is the same to relate the concept of \textit{utility} with the concept \textit{meaning}.

In addition, another separate .owl file for the spatial hierarchy has been created. In this file the real objects and real rooms are stored. Figure \ref{ontoKR3} shows the graph where entities are related with instances. In the experiments two rooms have been taken into account. One corresponds to a kitchen and the other to an office. Room 1 (the office) contains two chairs and a computer. Instead, Room 2 contains nothing.


The KRReasoner class inherits from the Reasoner class, in the same way that the BDReasoner class does. Therefore, both classes have the same methods. The difference is that KRReasoner is implemented by accessing the ontology contained in a .owl file through a server that works with prolog.

Three different query architectures have been used but all request the execution of different prolog instructions on the server.

\section{EXPERIMENTAL RESULTS}

The reasoning system has been implemented with two different technologies. One of them is based on access to a relational database that has been expressly designed to store the necessary concepts that describe an environment, its entities and their relationships. The other is based on the KnowRob system that researchers use at the Institute of Artificial Intelligence at the University of Bremen, Germany.

The integration of both systems of reasoning is simple because both reasoners inherit from a superclass that defines the way in which the rest of the navigation system interacts with the reasoner. The tests that have been carried out have the objective of verifying on the one hand that the information extracted from both systems of reasoning is the same and on the other hand, making an efficiency comparison.

A program that performs a battery of tests on both reasoners has been created to test the reasoning system implemented with KnowRob and to compare it with the reasoner implemented with databases. The tests consist of testing each of the methods offered by the reasoner. The methods are a way of consulting (or inferring) information. The requested information is entered by the user who is performing the tests. Although in the real system (when the reasoner is integrated in a robotic platform) they are the planner or the explorer. The reasoner in a mobile robot interacts with the planner or explorer. But the queries have to be invoked by the user to perform these tests. This allows the user to test each method. In addition, this test program allows to activate the reasoning with KnowRob and with the system based on the relational database indistinctly, as well as allowing the user to indicate the number of times that each operation should be repeated. This is because a parameter for queries to be made a certain number of times has been added. The execution time that the test program returns is the average of times of all the times the query is requested. This eliminates the possibility that by chance a query will take a much longer or shorter time than it usually takes. In these tests a parameter of 100 executions was established. The ontology for the KRReasoner was created with Protege version 5.1. The query server in Prolog uses the version of swi-prolog 6.6.

\begin{figure} [ht!]
	\centering
	\includegraphics[width=\columnwidth]{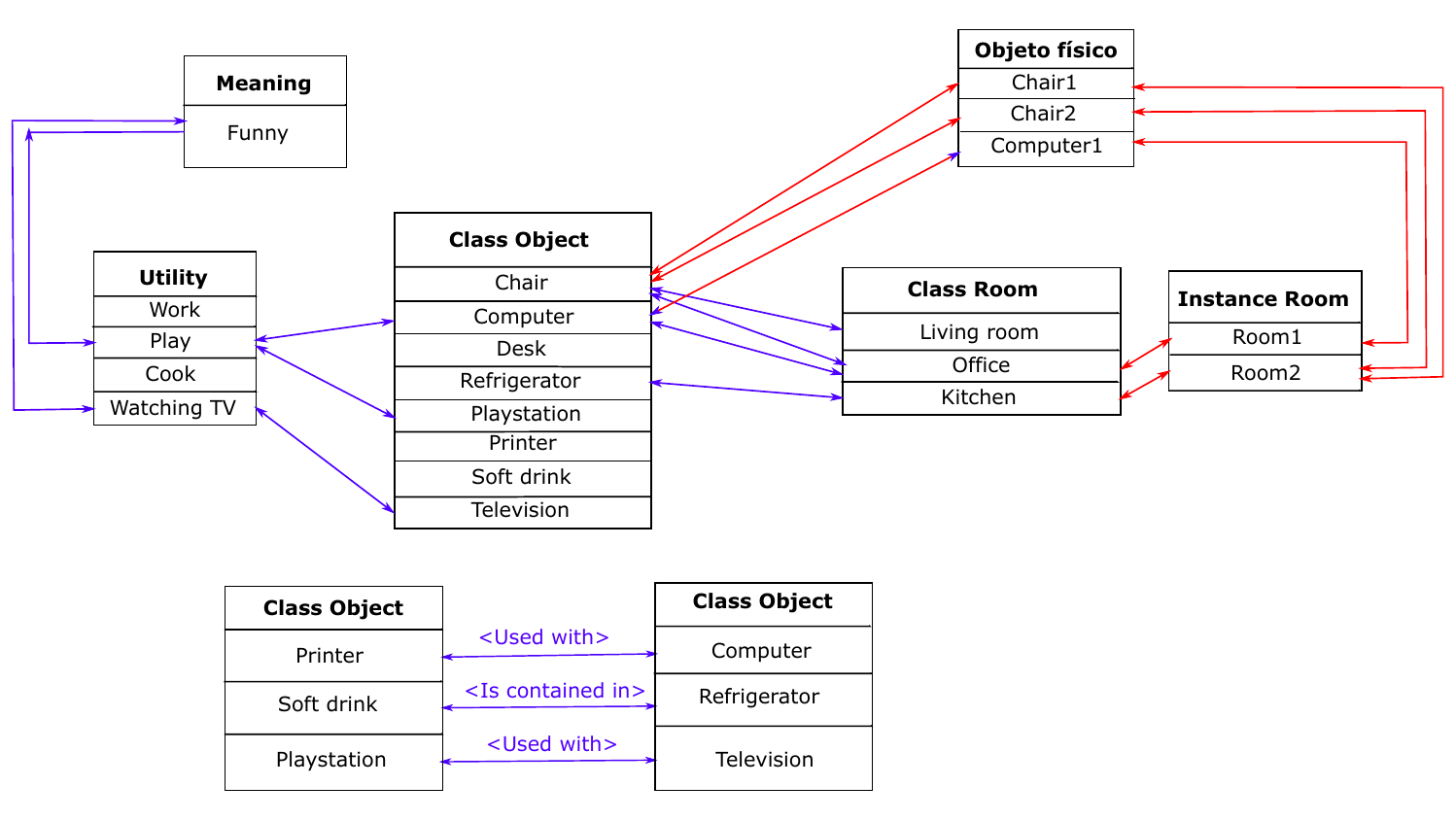}\\
	\caption{Knowledge on which the comparison experiment between reasoners is carried out. The red lines correspond to data about the real world and the purple lines to conceptual relations.}
	\label{expeKR}
\end{figure}

The information contained in both reasoners is shown in the figure \ref{expeKR}


\subsection{Performance Results}

The first question that this comparative wants to answer is whether both systems obtain the same conclusions. That is, if the reasoners make the same inferences with the same information. If so, the next interesting thing is to compare the execution times. The test program allows the user to execute each method a set of times with an input that is entered by keyboard. The program sends the input to each method of each selected reasoner and saves the execution times in a file.

\begin{table*}
	\centering
	\begin{tabular}{|c|c|c|c|c|c|}
		\cline{3-6}
		\multicolumn{2}{c}{} & \multicolumn{2}{|c|}{\textbf{Output}} &  \multicolumn{2}{|c|}{\textbf{Average execution time}}\\
		\hline
		\textbf{Method} & \textbf{Input} & \textbf{OutputDB} & \textbf{OutputKR} & \textbf{Time using DB} & \textbf{Time using KR}\\
		
		\hline
		Semantic labeling by object & Computer & Office & Office & 4 ms & 17 ms\\
		\hline
		Obtain Class Room from Instance Room & Room2 & Kitchen & Kitchen & 3 ms & 20 ms\\
		\hline
		Obtain Class rooms containing an object & Chair & Living room & Living Room & 2 ms & 29 ms\\
		&       & Office & Office & & \\
		\hline
		Obtain conceptually related objects & Soft drink & Refrigerator & Regrigerator& 3 ms & 50 ms\\
		& Printer & Computer & Computer &  & \\
		\hline
		Objects that serve a specific utility         & Work & Computer & Computer & 3 ms & 16 ms\\
		\hline
		Objects whose use has a meaning associated & Funny & Playstation & Computer & 2 ms & 63 ms\\
		&           & Television & Playstation&  & \\
		&           & Computer & Television &  & \\
		\hline
		Probable location of an object & Soft drink  & Kitchen (R) & Kitchen (R) & 4 ms & 49 ms\\
		\hline
		Physical room that fits conceptual room & Office & Room1 & Room1& 3 ms & 19 ms\\
		\hline
		Objects contained in physical room & Room1 & Chair & Chair&  3 ms & 21 ms\\
		&       & Computer & Computer &  & \\
		\hline
		Physical objects that fit with conceptual object   & Chair & Chair1 & Chair1 & 2 ms & 20 ms\\
		&       & Chair2 & Chair2& & \\
		\hline
		Conceptual name of a physical object  & Chair1 & Chair & Chair & 2 ms & 27 ms\\
		\hline
		Obtain all conceptual objects    &  ---   & List (*) & List (*) & 4 ms & 49 ms\\
		\hline
	    Obtain all actions / utilities     & --- & List (**) & List (**)& 2 ms & 32 ms\\

		\hline
	\end{tabular}
	\caption{Result of the comparison of the outputs between reasoners implemented with the two technologies.. (R): Refrigerator. (*) Complete list of conceptual objects, the order varies according to the reasoner. Each reasoner returns the 8 objects. (**) List of utilities, the order varies. Each reasoner returns 4 utilities.} \label{tablaKR1}
\end{table*}

Table \ref{tablaKR1} contains the information regarding the inputs that were sent to each method. It also shows the outputs that the reasoners returned. It is observed that both reasoners conclude the same, with the unique feature that when a query returns several entities as a response, the order of these entities does not have to coincide. This is normal because the ontology designed for KnowRob does not set any criteria for the ordering.

The average execution times are shown in table \ref{tablaKR1}. Each query was repeated 100 times. At first glance, the execution time of the system that accesses the relational database is considerably less than the execution time of the system based on KnowRob. The conclusion with the data taken and shown in the table is that on average of all the executions in the experiment, the system implemented with database take 8.9\% of the time that a query need with the system based on KnowRob. Which implies that the system based on the database is 11.13 times faster. This allows the machine's resources to be released and allows the performance of the semantic navigator to increase. It is also important to mention that the experiments have been carried out with very little volume of information. But the data suggests that the system based on KnowRob is more susceptible to their times increase when the queries are more complex or return more results. For example, in the case of the method of \textit{Objects whose use has associated a meaning} the average time goes up to 63 ms, the method for \textit{Semantically related objects} the time reaches 50 ms or the method for \textit{Probable location of an object} where 49 ms are reached. However, with the database-based system, those queries are barely heavier. This is because the databases are designed to work with a large volume of information.


The comparison between the two techniques has been carried out and a qualitative assessment is shown in Table \ref{tablaFinal}. This assessment is made on the basis of the user experience. The reasoning system of KnowRob is a powerful tool. It is suitable for any robot that requires reasoning about its environment. The reasoning system based on databases is an ideal solution for the specific task of navigation. This system offers faster execution and an easier way to define new environments.

\begin{table}[ht]
	\centering
	\resizebox{8.5cm}{!} {
	\begin{tabular}{|c|c|c|}
		\hline
		\textbf{Characteristic} & \textbf{Using KnowRob} & \textbf{Using database reasoner}\\
		
		\hline
		Scalability & Good & Complicated\\
		\hline
	    Execution Time & Higher & Lower\\
		\hline
	    Application for navigation & Good & Very Good\\
	    \hline
		Application for generic reasoning & Optimum & Possible\\

		\hline
	\end{tabular}
}
	\caption{Comparative of the implemented methods} \label{tablaFinal}
\end{table}

\subsection{Robot Navigation test}
\begin{figure} \centering
	\includegraphics[width=0.85\columnwidth]{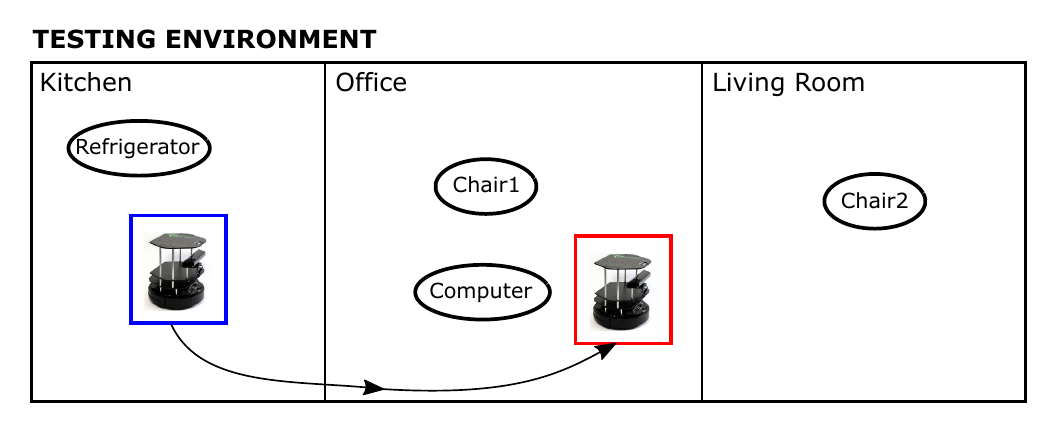}\\
	\caption{Testing environment}
	\label{testingRoom}
\end{figure}

Tests have been conducted on the turtlebot trying to check both reasoning systems. Figure \ref{testingRoom} is a schematic drawing of
the environment and represents a robot displacement from the kitchen to the office. The robot receives the requests compiled in table \ref{tablaKR1}. The system is forced to consider all the possibilities. To do this the user responds with negative showing dissatisfaction with the results. This causes the system to reasoning again. 

The Planner chains the answers of the reasoner. It is a recursive process. If the reasoner returns a different concept to a physical instance of an object or a room, the Planner again summon the reasoner with that concept. In these experiments there were defined only 5 physical instances of concepts, as shown in the Figure \ref{expeKR} and \ref{ontoKR3}. They are one kitchen room, one office room, two chair objects and a computer object. 

\begin{figure} \centering
	\includegraphics[width=1\columnwidth]{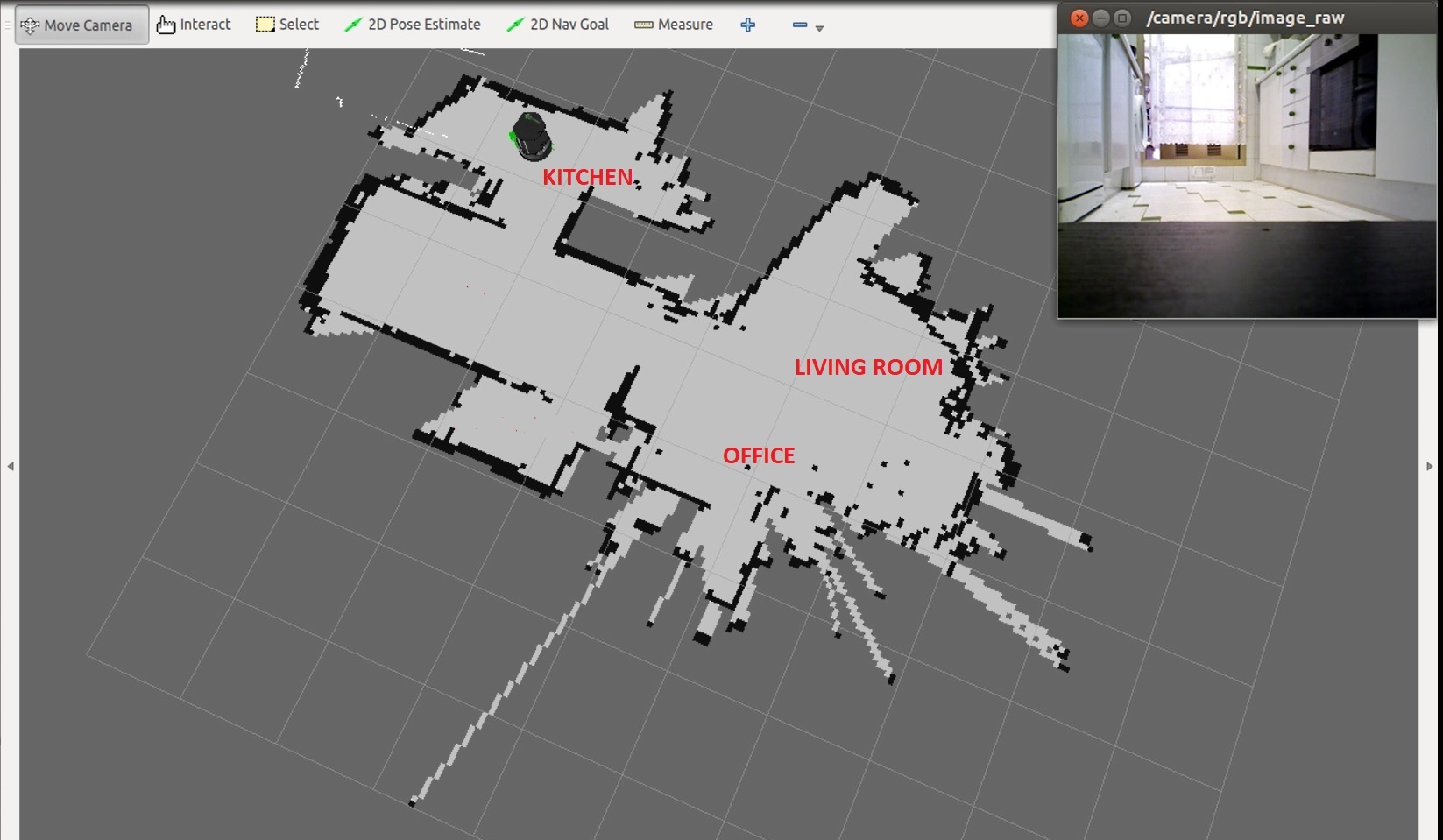}\\
	\caption{Geometric map of environment.}
	\label{figuraMapaVIP}
\end{figure}

The robot explored the environment using the algorithm included in ROS GMapping and obtained the map of Figure \ref{figuraMapaVIP}. Furthermore, the robot had the knowledge represented with the ontology described in figure \ref{expeKR}. In addition to the list of concepts shown in the figure, the system has more information. The computer has the utilities \textit{play} and \textit{work}. Television has the utility \textit{watching television}. The playstation has the \textit{play} utility. The printer is used with the computer. What is in the refrigerator has the characteristic \textit{cold}. The refrigerator contains the soft drink. The playstation is used with television.

\begin{figure} \centering
 \includegraphics[width=0.9\columnwidth]{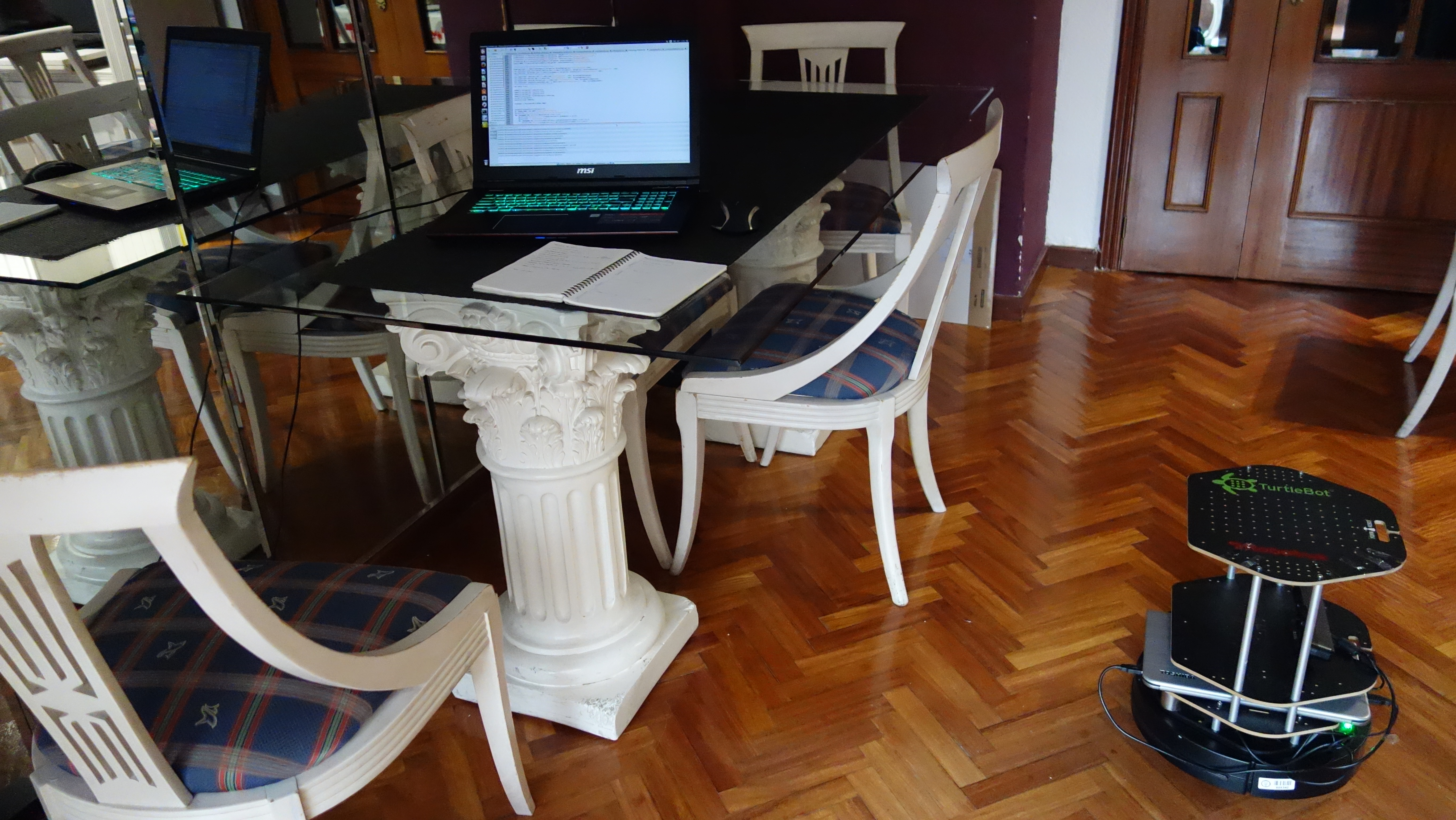}\\
 \caption{Robot in a place for working.}
 \label{enelordenador}
\end{figure}

\begin{figure} \centering
 \includegraphics[width=0.9\columnwidth]{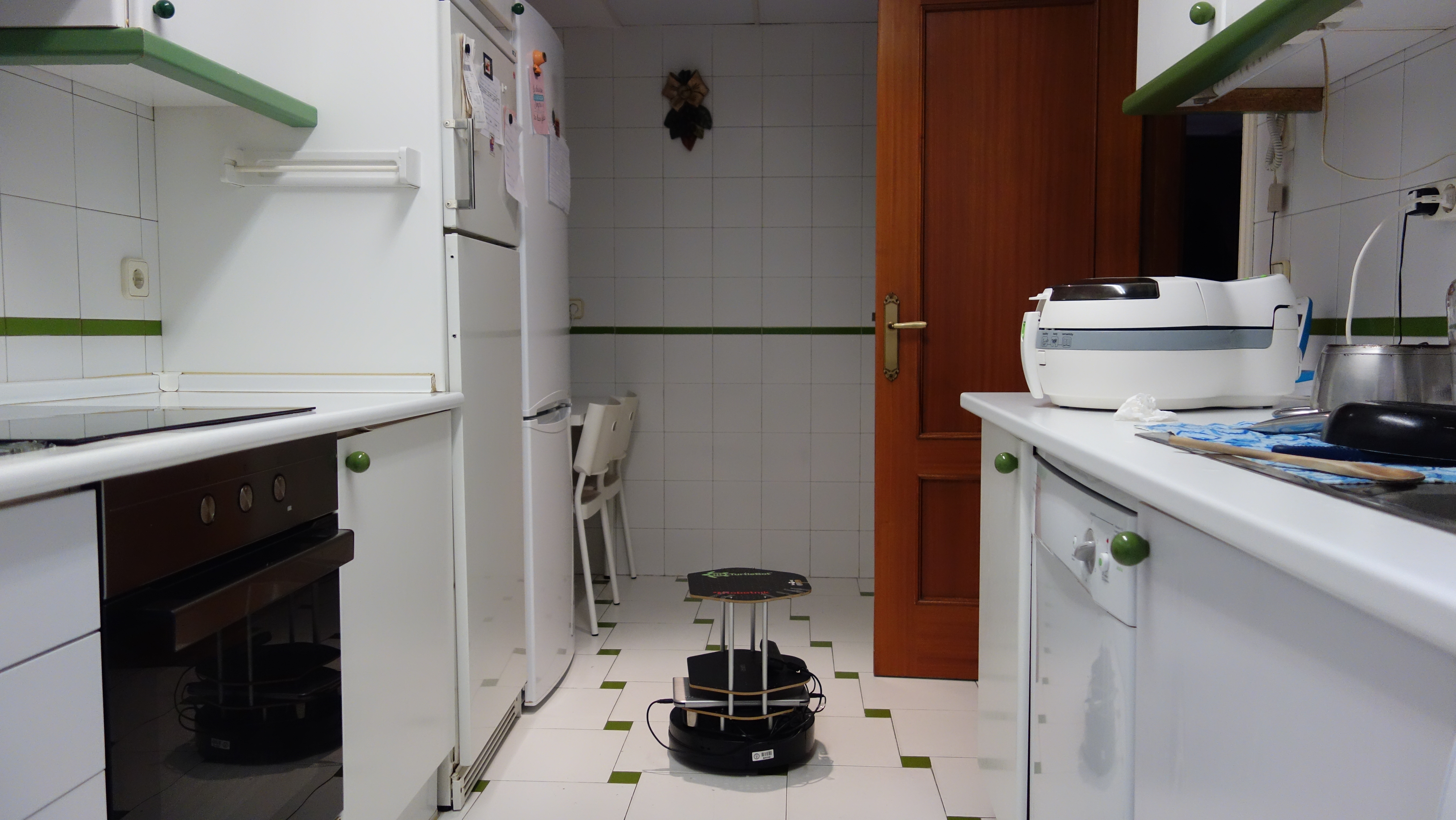}\\
 \caption{Final position when the robot searches for a softdrink.}
 \label{enlacocina}
\end{figure}

This set of experiments produced a series of displacements for the robot. Figures \ref{enelordenador} and \ref{enlacocina} show some of these experiments. Figure \ref{enelordenador} shows the destination that reached the robot  when the user asked to go to a site to work. In both images, the reasoning system based on KnowRob was being used. The robot went to the office because the computer is in the office and computer has the \textit{work} utility. Figure \ref{enlacocina} shows the destination reached by the robot when a soft drink was the request entered. The robot went to the kitchen because the refrigerator is in the kitchen and the refrigerator contains the soft drink. 

\section{CONCLUSIONS}

Although both systems are valid, the use of the database-based system has generated debates about its effectiveness and characteristics. The creation of the views and SQL queries that the inference system needs is not a simple or intuitive process but it is a one-time effort and this applies to any environment that is described with the same ontology represented in the base of relational data. Once the views are created, they are used for any environment that is entered in the tables of the database.

The inference system based on relational databases requires the information of the objects and their characteristics, types of room, actions performed by the objects and interactions are introduced in the tables. This gives an advantage over the other inference system implemented with KnowRob. With KnowRob, first defining an ontology is necessary. That ontology is generated in a .owl file. For this, knowledge of the Protege or similar tool is required. That ontology is only valid for a specific environment. In addition, the installation of Knowrob is not so simple nor the execution so light.

The system implemented with Knowrob has as an advantage a greater facility to scale the ontology. This is very useful for performing more complex tasks of cognitive robotics. However, for the problem of navigation, the system implemented with databases does not suffer from the difficulty of expanding the ontology since human environments conform to very specific patterns. The relations between objects, their function and the room in which they are found exist throughout the civilized world. However, in tasks more complex than navigation, the system based on KnowRob allows much greater adaptability.

Regarding the efficiency of the system, the reasoner implemented with the database wins the comparison. The execution times in the reasoning system that directly uses the database are considerably lower than the execution times of the reasoner that uses KnowRob's technology. The experiments of comparison between the reasoning system that uses databases and the reasoning system used by Knowrob have been carried out starting from the same ontological model, which has been represented in an equivalent way in the respective models, so they contain the same relationships between conceptual entities.

The proposed reasoners has been tested in a real robot, showing its functionality and performing semantic abstraction navigation tasks in a properly way.

\addtolength{\textheight}{-12cm}   

\section*{ACKNOWLEDGMENT}

The research leading to these results has received funding from the RoboCity2030-III-CM project (Rob\'otica aplicada a la mejora de la calidad de vida de los ciudadanos.
fase III; S2013/MIT-2748), funded by Programas de Actividades I+D en la Comunidad de Madrid and cofunded by Structural Funds of the EU and NAVEGASE-AUTOCOGNAV project (DPI2014-53525-C3-3-R), funded by Ministerio de Econom\'ia y Competitividad of SPAIN.

\bibliographystyle{ieeetr}
\bibliography{semanticNav}

\end{document}